# Image Dehazing using Bilinear Composition Loss Function

Hui Yang, Jinshan Pan, Qiong Yan, Wenxiu Sun, Jimmy Ren, Yu-Wing Tai

SenseTime Group Limited

## Abstract

*In this paper, we introduce a bilinear composition loss function to address the problem of image dehazing. Previous methods in image dehazing use a two-stage approach which first estimate the transmission map followed by clear image estimation. The drawback of a two-stage method is that it tends to boost local image artifacts such as noise, aliasing and blocking. This is especially the case for heavy haze images captured with a low quality device. Our method is based on convolutional neural networks. Unique in our method is the bilinear composition loss function which directly model the correlations between transmission map, clear image, and atmospheric light. This allows errors to be back-propagated to each sub-network concurrently, while maintaining the composition constraint to avoid overfitting of each sub-network. We evaluate the effectiveness of our proposed method using both synthetic and real world examples. Extensive experiments show that our method outperforms state-of-the-art methods especially for haze images with severe noise level and compressions.*

## 1. Introduction

In outdoor photographies, because of the presence of fog, dust, mist, and fumes, images captured under such environments usually have low contrast, color shift and poor visibility. This effect may be an annoyance to amateur, commercial, and artistic photographers as well as undermine the quality of underwater and aerial photography. The deterioration of image quality can also degrades the performance of algorithms in many computer vision tasks such as detection and tracking. Therefore, it is important to remove haze as part of imaging post-processing pipeline.

Previous studies in image dehazing formulate the problem as an image composition problem [16, 20, 5, 22, 8, 9, 7, 15, 1], where the effect of haze can be modeled as:

$$\mathbf{I}(x) = \mathbf{J}(x)\mathbf{T}(x) + \mathbf{A}(1 - \mathbf{T}(x)), \quad (1)$$

where $\mathbf{I}(x)$ is the haze image, $\mathbf{J}(x)$ is the scene radiance, $\mathbf{A}$ is the atmospheric light, and $\mathbf{T}(x) \in (0, 1]$ is the medium transmission map. The transmission map describes the portion of light that reaches to the camera without scattered and it may be modeled as $\mathbf{T}(x) = \exp^{-\beta d(x)}$ where $\beta$ is the medium extinction coefficient and $d(x)$ is the scene depth. The goal of image dehazing is to estimate $\mathbf{J}(x)$ (with $\mathbf{T}(x)$ and $\mathbf{A}$ as by-products) from a single input image $\mathbf{I}(x)$. By Equation (1), image dehazing is a highly ill-posed problem since multiple solutions exist.

Numerous methods [17, 20, 5, 22, 8, 9, 15, 1, 23] have been proposed to solve the image dehazing problem. The dark channel prior [8, 9] is one of the most widely used prior in solving the image dehazing problem. It assumes the value of one of the three color channels in local region is closed to zero. This assumption, however, is invalid for sky regions or objects with white/light color. Also, this prior usually over-estimate the amount of transmission, therefore, over-enhance the local contrast of $\mathbf{J}(x)$ and the estimated $\mathbf{J}(x)$ usually has color shift artifacts. Another problem of Equation (1) is that it ignores the effects of image noise and compression. Although assuming $\mathbf{T}(x)$ is piecewise smooth and $\mathbf{A}$ is globally constant can make the problem tractable, the effects of image noise will be amplified by a factor of $1/\mathbf{T}(x)$ especially for the methods that take a two-stage approach which first estimate $t(x)$ and $\mathbf{A}$, followed by clear image estimation by $\mathbf{J}(x) = \frac{\mathbf{I}(x) - \mathbf{A}(1 - \mathbf{T}(x))}{\mathbf{T}(x)}$.

To include the effects of image noise, we modify Equation (1), and propose the following model:

$$\begin{aligned}\mathbf{I}(x) &= \mathbf{J}(x)\mathbf{T}(x) + \mathbf{A}(1 - \mathbf{T}(x)) + N(x), \\ &= \mathbf{J}(x)\mathbf{T}(x) + (\mathbf{A} + N'(x))(1 - \mathbf{T}(x)), \\ &= \mathbf{J}(x)\mathbf{T}(x) + \mathbf{A}'(x)(1 - \mathbf{T}(x)),\end{aligned} \quad (2)$$

where $N'(x) \approx \frac{N(x)}{(1-\mathbf{T}(x))}$ and $\mathbf{A}'(x) \approx \mathbf{A} + N'(x)$. Compare with Equation (1), we convert the globally constant, $\mathbf{A}$, into a spatially varying $\mathbf{A}'$ which absorbs the effects of image noise and other unmodeled artifacts in Equation (1).

In order to solve Equation (2), we utilize a deep learning approach. In other words, we propose to use a data driven approach to learn a deep convolutional neural network to estimate $\mathbf{T}(x)$, $\mathbf{A}'(x)$ and $\mathbf{J}(x)$ from a single input image $\mathbf{I}(x)$. We utilize a bilinear network to simuta-neously estimate $\mathbf{T}(x)$ and $\mathbf{A}'(x)$, followed by the estimation of $\mathbf{J}(x)$. Since $t(x)$, $\mathbf{A}'(x)$ and $\mathbf{J}(x)$ are correlated with each other, we introduce the bilinear composition loss function derived from Equation (2) to back propagate errors to each sub-network concurrently in order to avoid overfitting caused by individual training of each sub-network. Following the piecewise smoothness assumption of $\mathbf{T}(x)$, we include the guided filtering [10] to post-process the output of transmission network. During the back propagation, the composition errors have also gone through the same guided filtering where the piecewise smooth errors are back propagated to the transmission network, and the residual high frequency errors are back propagated to the atmospheric network. This allows us to estimate a more accurate transmission map and atmospheric map where high frequency image noise/compression artifacts can be well-separated into the atmospheric map. After obtaining the transmission and atmospheric maps from the bilinear network, we solve Equation (2) to obtain $\mathbf{J}(x)$. To this end, we propose to use an image enhancement net to further enhance the estimation of $\mathbf{J}(x)$ in order to achieve a high quality dehazed image.

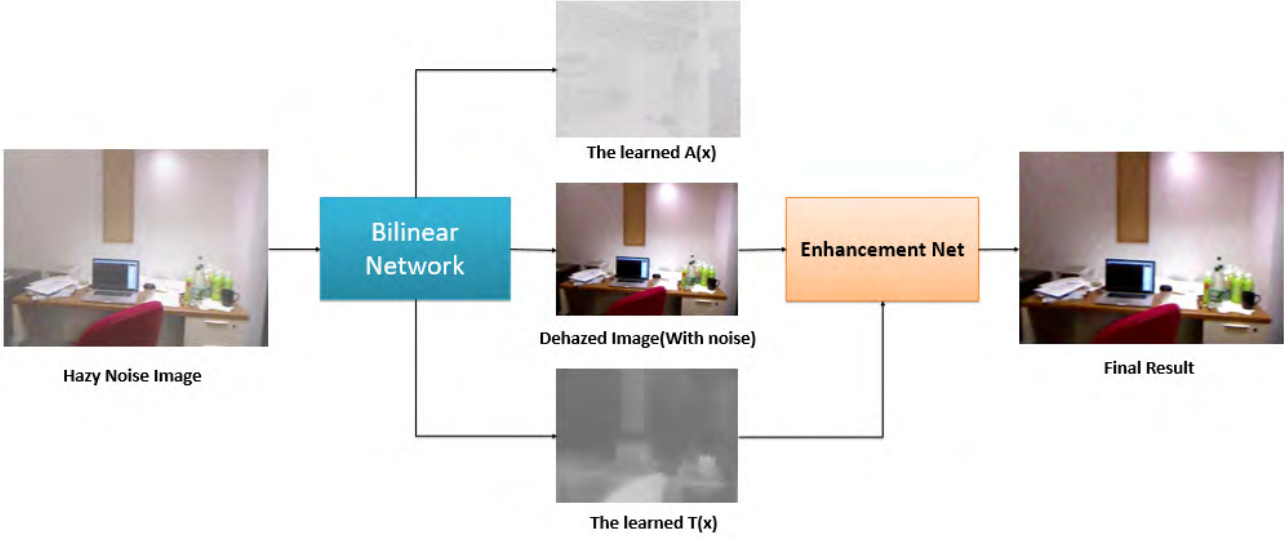

Figure 1: Overview of our method.

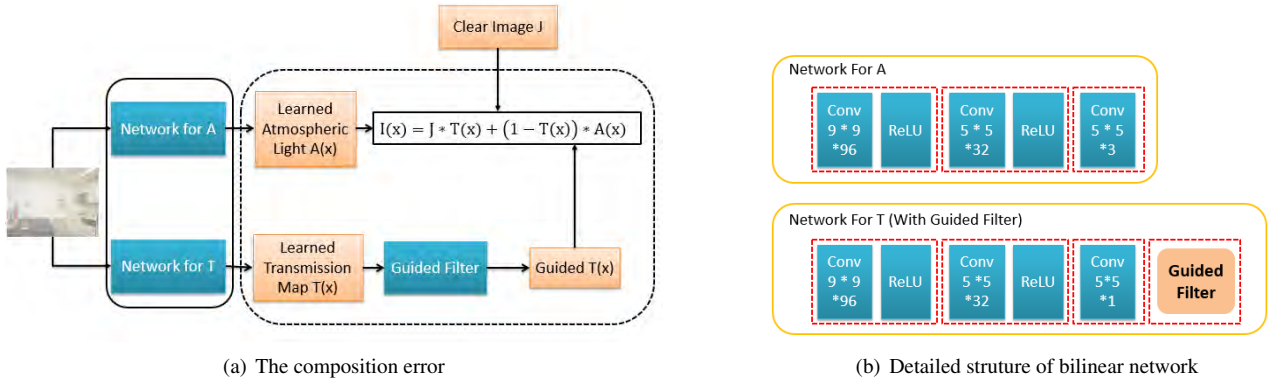

(a) The composition error

(b) Detailed struture of bilinear network

Figure 2: Overview of our method and the bilinear network architecture.

## 2. Related Work

Early works in image dehazing often require multiple inputs, such as additional images captured at different time/spectrum/polarization, depthmap, or 3D models. Representative works include [11, 24, 19, 16, 20, 14]. However, it is difficult to obtain these inputs without resolving additional devices or manual inputs. In most cases, there is only a single hazy image.

Recent works in single image dehazing often require to define some image priors. Fattal [5] assumes a hazy image can be separated into regions of constant albedo, and then infer the scene transmission based on the constant albedo assumption. Tan [22] proposes to maximize local contrast to enhance image visibility. He *et al*. [8, 9] introduces the dark channel prior. Fattal [6] proposes the color-line prior. A variety of multi-scale haze relevant features are reviewed and analyzed by Tang *et al*. [23]. Most recently, Ren *et al*. [18] propose to use multi-scale convolutional neural networks to estimate transmission map directly from a hazy image.

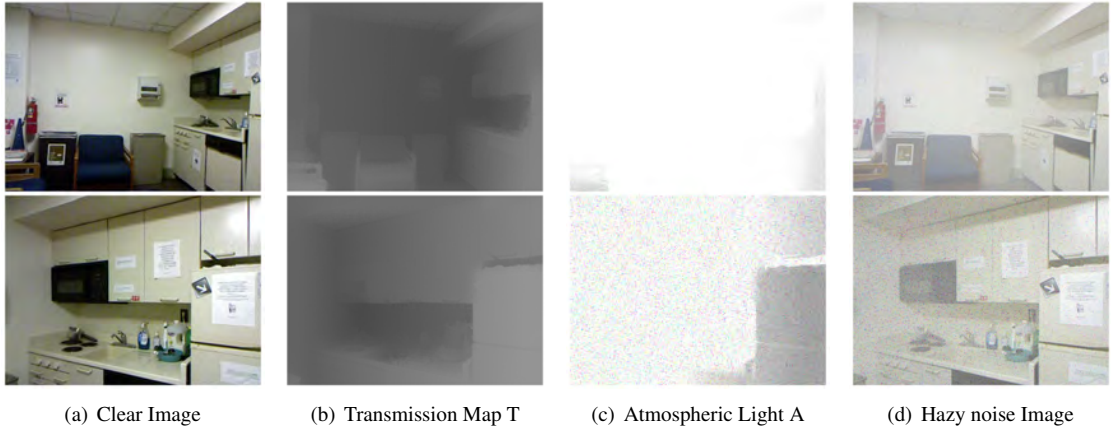

Figure 3: (a), (b) and (c) are samples showing clear images, transmission maps, noise A and synthesized hazy noise images in the training dataset.

Despite the good results reported in the aforementioned methods, one practial issue that have been largely overlooked is the problem of image noise and compression. As discussed before, the two-stage approach tends to amplify image noise when the transmission map and atmospheric map are assumed to be smooth/piecewise smooth. Li *et al*. [12] proposes to suppress these artifacts using spatially varying local regularization. Chen *et al*. [3] propose the gradient residual minimization to jointly recovering haze-free image while minimizing the visual artifacts.

Compare with previous works, ours also utilizes deep learning to estimate transmission map directly from a single hazy image. Different from the previous works, we also simultaneously estimate the atmospheric map which separate image artifacts before haze-free image estimation. Thus, we can achieve a much better haze-free image after our bilinear network. The image enhancement network serves as a supplement to remove residual artifacts and further enhance our dehazed images.

## 3. Algorithm

An overview of our method is presented in Figure 2. Our method takes a single hazy image, and use a bilinear network to learn the transmission map and atmospheric map simultaneously, followed by clear image estimation using an image enhancement net. The two branchs of the bilinear network shares the same bilinear composition loss. In the following, we will explain the details of each component of method.

**Bilinear Network**

The architecture of our bilinear CNN is presented in Figure 2. It consists of two branches, each branch has three convolutional layers. In the sub-branch of **T**-network, we include a guided filtering to enforce piecewise smoothness of **T**. Note that the back-propagated errors through the **T**-network has also gone through the guided filtering. We will prove its correctness later in this section.

**Bilinear Composition Loss Function**

We define the bilinear composition loss function as follow:

$$\mathcal{L}_c = \frac{1}{N} \sum_{i=1}^{N} \|(\mathbf{J}_i * \mathbf{T}_i + \mathbf{A}_i * (1 - \mathbf{T}_i)) - \mathbf{I}_i\|^2 \qquad (3)$$

where $N$ is total number of training examples, **A** and **T** are the outputs of the bilinear network, **J** is the ground truth clear image, and **I** is the input hazy image. The coordinate, $(x)$, in Equation (3) is omitted for simplicity of representation. The bilinear composition loss function measures the least square errors of Equation (2), where **A** and **T** denote the atmospheric map and transmission map respectively.

To back-propagate errors to each sub-branch of the bilinear network, we compute the partial derivative of Equa-

tion (3) with respect to **A** and **T**:

$$\frac{\partial \mathcal{L}_c}{\partial \mathbf{A}} = \frac{1}{N}\sum_{i=1}^{N}(1-\mathbf{T}_i)*\mathcal{E}_i, \quad (4)$$

$$\frac{\partial \mathcal{L}_c}{\partial \mathbf{T}} = \frac{1}{N}\sum_{i=1}^{N}(J-\mathbf{A}_i)*\mathcal{E}_i, \quad (5)$$

$$\mathcal{E}_i = (\mathbf{J}_i*\mathbf{T}_i + \mathbf{A}_i*(1-\mathbf{T}_i)) - \mathbf{I}_i. \quad (6)$$

In order to regularize the network for better convergence, we use the ground-truth clear image together with the estimated **A** and **T** to measure the composition loss. Additionally, we have included a regularization to each sub-branch which minimizes the euclidean distance between the estimated **A** and **T** and the ground truth **A** and **T** respectively. Namely, the back propagation for A and T become:

$$\frac{\partial \mathcal{L}_c}{\partial \mathbf{A}} = \lambda_1 * \frac{1}{N}\sum_{i=1}^{N}(1-\mathbf{T}_i)*\mathcal{E}_i + \lambda_2 * \frac{1}{N}\sum_{i=1}^{N}(\mathbf{A}_i-\mathbf{A}_{GT}), \quad (7)$$

$$\frac{\partial \mathcal{L}_c}{\partial \mathbf{T}} = \lambda_1 * \frac{1}{N}\sum_{i=1}^{N}(J-\mathbf{A}_i)*\mathcal{E}_i + \lambda_2 * \frac{1}{N}\sum_{i=1}^{N}(\mathbf{T}_i-\mathbf{T}_{GT}). \quad (8)$$

**Back propagation with Guided Filter** In our bilinear network, we include the guided filter to post-process the output of **T**-network. Suppose $t$ is the output of the convolutional neural network, the process of guided filtering can be modelled by the following equation:

$$\mathbf{T} = G(I,t) \quad (9)$$

where $G$ denotes the guided filter operation, and $I$ is the hazy image to guide the filtering process. Since $I$ is fixed, we can re-write Equation (9) into a matrix-vector form [4]:

$$\mathbf{T} = W^I t, \quad (10)$$

where $W^I$ denote the matrix form of guided filter. The value of the $i,j$-th entry of $W^I$ is equal to

$$W^I_{ij} = \frac{\sum_{k\in\Omega_i\cap\Omega_j}\left(1+\frac{(I_i-E_{\Omega_k}(I))(I_j-E_{\Omega_k}(I))}{D_{\Omega_k}(I)+\epsilon}\right)}{(\sum_{k\in\Omega_i}1)(\sum_{k\in\Omega_j}1)}, \quad (11)$$

where $E_{\Omega_i}(I)$ and $D_{\Omega_i}(I)$ denote the arithmetical mean and variance of $I$ in the local support region defined by $\Omega_i$. Note that the $W^I_{ij}$ depends only on the signal of guided image, and it is independent to the signal to be filtered.

To back-propagate errors through the guided filter, suppose $\partial\mathcal{E}/\partial\mathbf{T}$ be the gradient of the loss functions before the guided filter, by chain rule of gradients we have:

$$\frac{\partial G(\mathcal{E})}{\partial \mathbf{T}} = \frac{\partial W^I\mathcal{E}}{\partial \mathbf{T}} = W^I\frac{\partial \mathcal{E}}{\partial \mathbf{T}} = G(\frac{\partial \mathcal{E}}{\partial \mathbf{T}}). \quad (12)$$

Thus, we can consider the guided filtering as a guided filter layer, and its forward and backward passes are defined by Equation (9) and Equation (12) respectively.

**Image Enhancement Net**
After we obtain the atmospheric map and the transmission map, the dehazed image can be solved in closed form. However, we note that the dehazed image may still contain small amount of remaining noise although most noise/compression artifacts have been separated in the atmospheric map. We adopt an enhancement net to exclusively learn the remaining noise from dehazed image with clear image being the ground truth target. Since the three-layer network structure has been proved of good capacity in learning A and T in the bilinear network, the same structure is applied to enhancement net. In order to accelerate the denoising process, an additional channel of learned T is added to the input. Smooth and almost noise-free, the learned transmission map guides the dehazed image with little noise towards the clean ground truth.

**Parameter Settings**
The three convolutional layers are comprised of filters varying in kernel sizes and numbers, as shown in Figure 2(b). The first layer has 96 filters of size 9*9 that takes the 3 channel hazy image as input. The second and third layer both consists of filters with size 5*5. The second layer maps the feature sets generated by the first layer to 32 channels, and the third convolutional layer interacts with the last loss layer by combing the 32 channels into feature maps. Unlike previous methods which treat A as one channel, we devise A to be three channels in order to better accommodate noise. For T, a single channel feature map is output. The ReLU units following the first two convolutional layers are of classical implementation. Zero padding is applied to convolutional layer, so the difference calculated in the loss layer is based on the final size of the output. Learning for t and A employs stochastic gradient descent (SGD) with 0.9 momentum for training. The learning rate is set to 0.001 for the first two convolutional layers and 0.0001 for the last one.

The value for $\lambda_1$ and $\lambda_2$ in the loss function are set as 0.1 and 0.9 respectively. $\lambda_2$ represents the updating speed of A and T learned from their own ground truth. $\lambda_1$ sets the weight for learning from composition error which avoids overfitting training.However $\lambda_1$ should be much smaller than $\lambda_2$ in case the composition error interferes with individual learning of A and T, since the two elements are both far from ground truth in the first few iterations.

## 4. Experimental Results

We evaluate the performance of our proposed method in this section. We implement our bilinear network using MatConvNet. The whole experiment is conducted on a laptop with an Intel Core i7 CPU and NVIDIA GM204 [GeForce

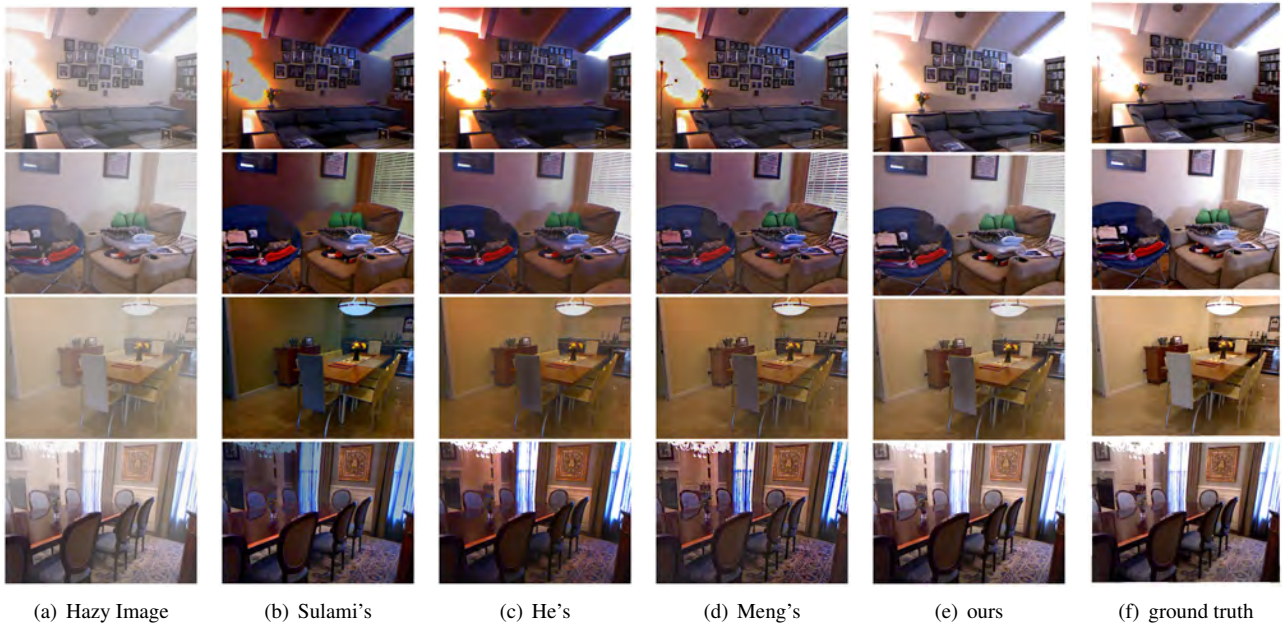

| (a) Hazy Image | (b) Sulami's | (c) He's | (d) Meng's | (e) ours | (f) ground truth |

Figure 4: Dehazed results recovered from synthetic indoor noise-free images.

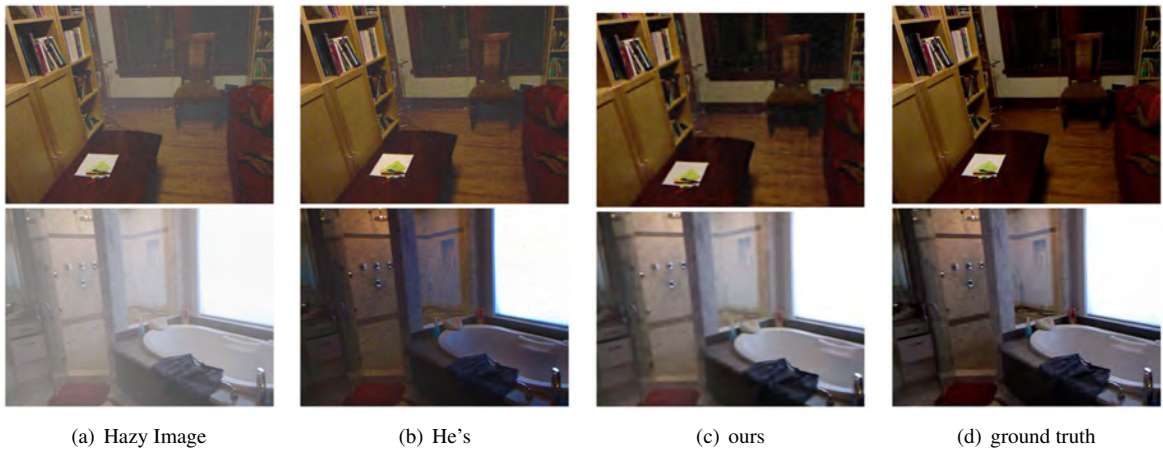

| (a) Hazy Image | (b) He's | (c) ours | (d) ground truth |

Figure 5: Dehazed results recovered from synthetic indoor noise images.

GTX 970] GPU on Linux system. We will first describe our training process, followed by quantitative evaluation using synthetic examples, and qualitative evaluation on real world examples.

### 4.1. Training

The training data are composed of synthesized noise hazy image utilizing the NYU2 depth dataset [21]. We use the haze image model presented in Equation (2) to generate noise hazy image. The atmospheric light A is randomly generated from values within (0.7, 1.0) and the thickness of haze is diversified by randomly sampling $\beta$ within (0.5,1.5). After generate a noise-free hazy image, guassian noise with zero mean and variance with in (0.003,0.01), salt and pepper noise with density within (0.015,0.01) is added to A, thus introduced to the synthesized hazy image. The training is conducted on patches with size $32 * 32$ sampled from a $480*640$ image. There are 600 images in the NYU2 dataset. In total, we sampled more than 200, 000 patches to train the networks. Figure 3 shows some of the training samples.

### 4.2. Evaluation and comparison

Our proposed bilinear convolutional network is first evaluated by testing its performnace on noise free image.The results are compared with some state-of-the-art methods. He's method [8] relies on the dark channel prior while Ren

| Average Metris | [8] | [15] | [18] | Ours |
|---|---|---|---|---|
| PSNR | 15.86 | 15.06 | 18.38 | 23.31 |

Table 1: Average PSNR of dehazed results on the noise-free synthetic indoor dataset.

| Image Size | [15] | [8] | [18] | Ours |
|---|---|---|---|---|
| 620*460 | 2.88 | 0.63 | 1.68 | 1.13 |

Table 2: Average run time (in seconds) on test images.

*et al.*'s method [18] is more recently advanced convolutional network-based solution. The validity of the bilinear CNN is verified using the Peak Signal-to-Noise Ratio (PSNR) metric. Then we use the synthesized noise hazy images to evaluate the dehazed results. Both experiments are conducted on two categories of image sets: indoor hazy images from the part of NYU2 dataset that are not used as training data and real world outdoor images collected from both the benchmark database and Internet.

**Synthetic dataset**
**Dehazed results of the noise free images** The dehazed effects of synthetic indoor hazy images are shown in Figure 4. Given the priors that each method utilizes to approximate the atmospheric light **A**, traditional approaches overestimate the transmission map, resulting in darker dehazed outputs. This is more evident in the case of white objects, such as the white wall and chair in the input hazy image. The proposed bilinear network can automatically learn **A** and **T** from the training data. The convolutional structure and the multiple non-linear mappings enable CNN to capture a wide range of haze degrees and presents a more sensible transmission map. Table 1 presents the PSNR values for each image recovered by various methods. Our bilinear network achieves the highest PSNR score on average, in accordance to the more pleasing visual results. This advantage is also manifested in the case of natural outdoor images, which are generally the major scenes that need to be dehazed.

**Dehazed results of noise images** The dehazed results of synthetic indoor hazy images are shown in Figure 5. Since most dehazing methods tend to boost noise during the contrast enhancement process, we only select He's results as representative for comparison. Our method can effectively dehaze hazy image with moderate noise compared to He's as shown by Figure 5.

**Effects of guided filter layer** The guided filter layer added at the back of the network for transmission map **T** smooths the learned **T** while preserves the edge information at the same time. Without the guided filter layer, the convolutional neural network tends to boost noise and generates halo effects as training continues. The dehazed results with and without the guided filter layer are shown in Figure 6.

**Effects of image enhancement net**
The haze-free image output by the bilinear network still has minor noise. This noise is transformed by the proposed bilinear network since hazy images, along with noise, are convolved together through the whole structure. This minor artifacts thus can no longer be addressed by simple denoise algorithm. The image enhancement net becomes necessary to reconstruct noise-free and haze-free image to approximate ground truth. Figure 7 demonstrates the dehazed results with and without the enhancement net. The remaining noise can be effectively removed while the contrast and dehazed effects are maintained.

**Real world examples**
**Dehazed results of the noise free images** For real world outdoor cases, we demonstrate dehazed results of the proposed network using six noise free scenes shown in Figure 8. For the sky region, our bilinear network restores the haze free image in a more visually pleasing way. The atmospheric light **A** is well approximated in respect to most natural scenes.

**Dehazed results of the noise hazy images** We compare our bilinear network with He's dark channel prior method using two outdoor scenes with moderate Gaussian noise added. The results are shown in Figure 9. He's results are not robust to noise. The artifacts are boosted once dehazed and contrast enhanced. As demonstrated in Figure 9(c), our method smooths the noise while enlarges color contrast for dehazing.

### 4.3. Runtime Comparison

The bilinear network runs relatively fast compared to some state of the art methods. Four images of size 620*460 in Figure 4 are used for evaluation. All compared methods are operated by their original codes posted on the project websites, which are implemented in MATLAB and are tested without GPU acceleration on a single machine. Table 2 shows the average runtime of each method dealing with the above images. Except for Hes, our network is the fastest to operate. This runtime can further be shortened if GPU acceleration is adopted.

## 5. Further Analysis and Discussions
### 5.1. The NYU2 dataset

Unlike [2] that produces artificial haze images by assigning randomly sampled transmission maps t to an image patch and assumes atmospheric **A** as value 1 for all the training data, we use authentic depth maps from NYU2 dataset to generate hazy images as shown in formula (1). Since the ground truth data is based on real value depths measured by

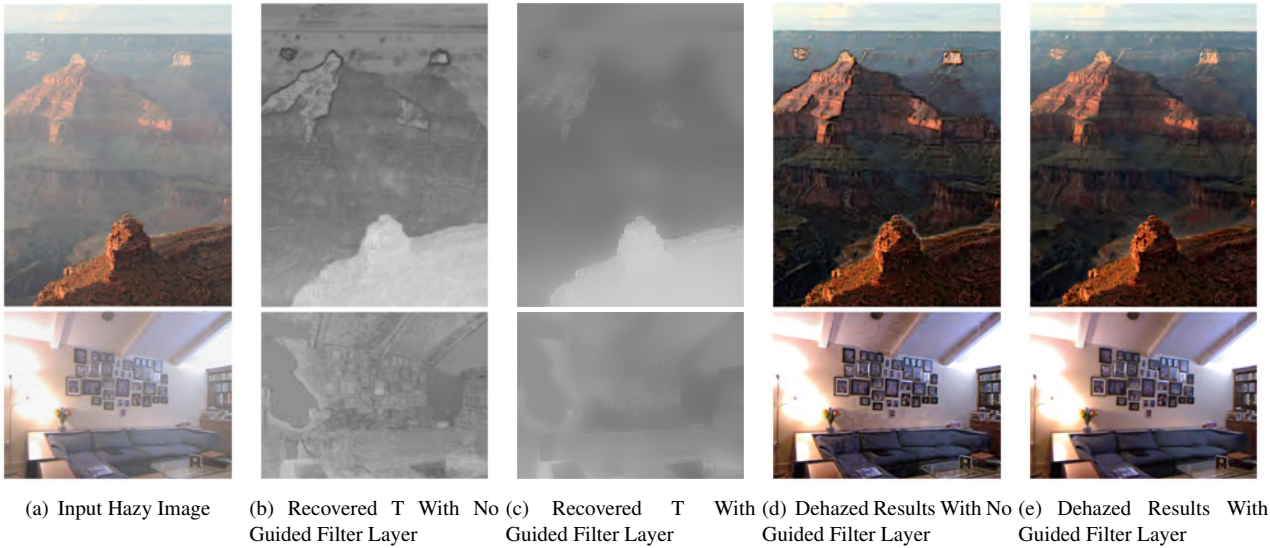

(a) Input Hazy Image  (b) Recovered T With No Guided Filter Layer  (c) Recovered T With Guided Filter Layer  (d) Dehazed Results With No Guided Filter Layer  (e) Dehazed Results With Guided Filter Layer

Figure 6: The effects of guided filter layer. Notice the halo effects without the guided filter.

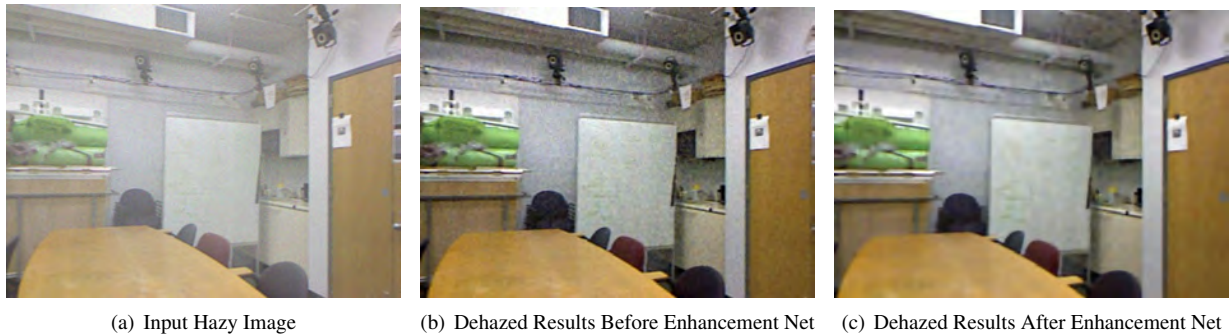

(a) Input Hazy Image  (b) Dehazed Results Before Enhancement Net  (c) Dehazed Results After Enhancement Net

Figure 7: The effects of enhancement net. Remaining noise are suppressed effectively.

Kinect camera , our training data represent intrinsic features of hazy images more reasonably. The Frida [13] dataset has been used to synthesize hazy images. It failed to achieve the same performance as NYU2 dataset as our experimental results suggest. Images from Frida database are basically computer-synthesized, so the structure of the contents are overall simple, which can hinder CNN network from understanding the internal depth map. Moreover, Frida dataset assumes the sky to be infinitely away, thus the ground truth depth maps for sky region are near 0. This leads to overestimation of **A** and results in darker dehazed images.

### 5.2. The value range of atmospheric light

The training value assigned to the atmospheric light **A** has a considerably big impact on the final recovered results. Setting a large value of **A** can lead to over-dehazed images while setting a small value of **A** leaves some haze unremoved. In our experiment, we found that initially setting **A** as 1 for all training data as in [2]. Though this achieves good results on testing data from the NYU2, real world images suffer from darker color in the haze-free regions. Randomly sampling **A** from [0.7,1.0] helps mitigating this problem. Moreover, learning **A** as a three-channel vector enables CNN stronger ability to capture artifacts of the hazy image.

## 6. Conclusion

In this paper, a bilinear composition loss function is presented to solve the image dehazing problem. We train the bilinear network using three errors: the composition error updated by the learned atmospheric light A and transmission map T simultaneously, the two branch errors calculated between these two elements and their ground truths. The bilinear network is proved to be effective to address the noise artifacts during dehazing. In order to remove the remaining noise, an enhancement net is attached to the back of the bilinear network to further polish the dehazed results. We compare our method with state-of-the-arts methods and obtain visually pleasing results both on the synthetic indoor and real world outdoor images. For the noise-free images, The advanced networks restore the atmospheric light **A** in a more natural way, especially for the sky region as shown in our experimental results. For hazy images with noise, the noise can be automatically separated into **A**. This operation naturally combines dehazing and denoising and achieves better results than traditional methods that leave noise unattended.

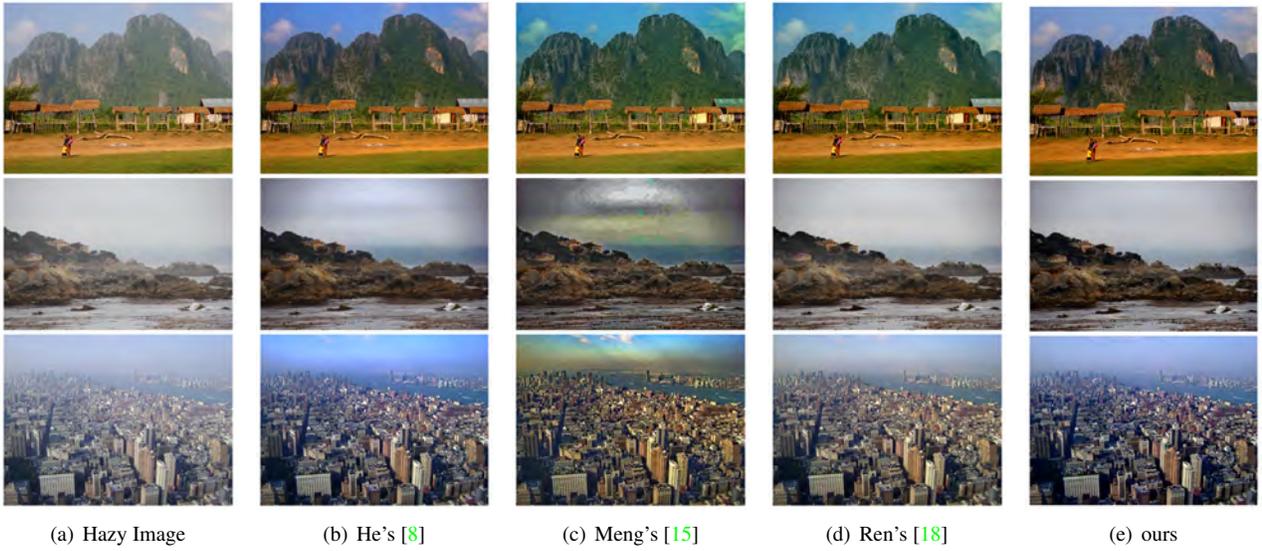

(a) Hazy Image    (b) He's [8]    (c) Meng's [15]    (d) Ren's [18]    (e) ours

Figure 8: Dehazed results recovered from noise free real world outdoor images.

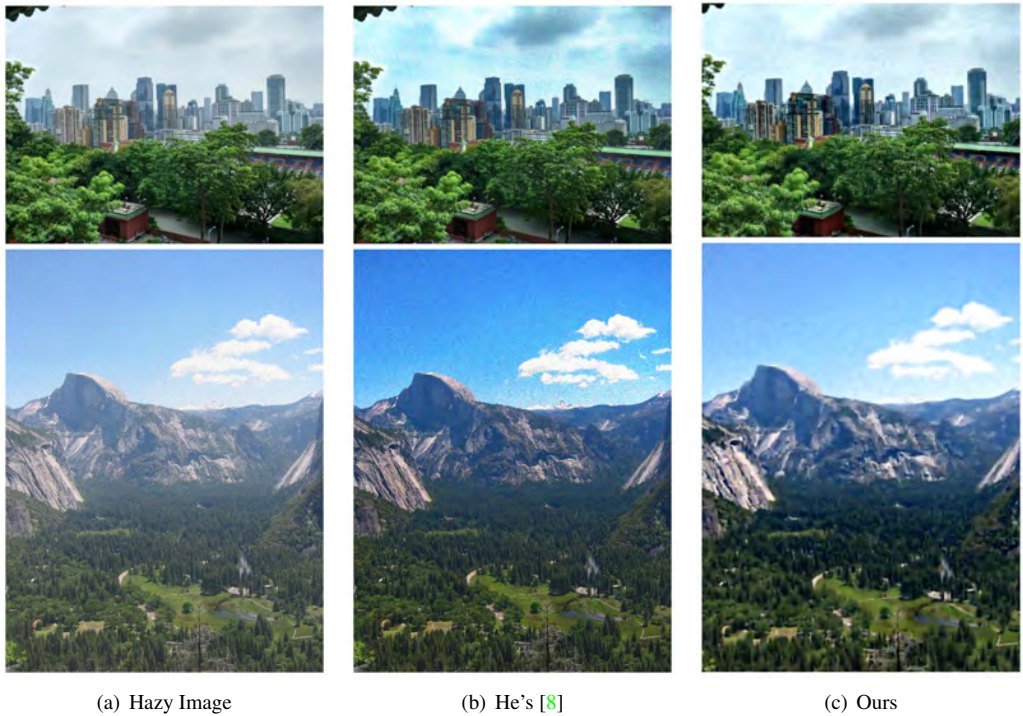

(a) Hazy Image    (b) He's [8]    (c) Ours

Figure 9: Dehazed results recovered from noised real world outdoor images.